\pdfoutput=1
\documentclass[11pt]{article}

\usepackage{acl}

\usepackage{times}
\usepackage{amsthm}
\usepackage{latexsym}
\usepackage{graphicx}
\usepackage{color}
\usepackage{caption}
\usepackage{subcaption}
\usepackage{bbm}
\usepackage{url}
\usepackage{amssymb}
\usepackage{amsmath}
\usepackage{multirow}
\usepackage{booktabs}
\usepackage{algorithm}
\usepackage{algpseudocode}
\usepackage{wrapfig}
\newtheorem{theorem}{Theorem}[section]

\newcommand{\algo}{\textsc{CeeBERT}}

\title{\algo{}: Cross-Domain Inference in Early Exit BERT}

\author{Divya Jyoti Bajpai \\
  Research Scholar \\
  IEOR Department \\
  IIT Bombay \\
  \texttt{divyajyoti.bajpai@iitb.ac.in} \\\And
  Manjesh Kumar Hanawal \\
  Associate Professor\\
  IEOR Department \\
  IIT Bombay \\
  \texttt{mhanawal@iitb.ac.in} \\}

\begin{document}
\maketitle

\begin{abstract}
%Through scaling up the model size and data volume, 
Pre-trained Language Models (PLMs), like BERT, with self-supervision objectives exhibit remarkable performance and generalization across various tasks. However, they suffer in inference latency due to their large size. To address this issue, side branches are attached at intermediate layers, enabling early inference of samples without requiring them to pass through all layers. However, the challenge is to decide which layer to infer and exit each sample so that the accuracy and latency are balanced. Moreover, the distribution of the samples to be inferred may differ from that used for training necessitating cross-domain adaptation. We propose an online learning algorithm named Cross-Domain Inference in Early Exit BERT (\algo{}) that dynamically determines early exits of samples based on the level of confidence at each exit point. \algo{} learns optimal thresholds from domain-specific confidence observed at intermediate layers on the fly, eliminating the need for labeled data. Experimental results on five distinct datasets with BERT and ALBERT models demonstrate \algo's ability to improve latency by reducing unnecessary computations with minimal drop in performance. By adapting to the threshold values, \algo{} can speed up the BERT/ALBERT models by $2\times$ - $3.5\times$ with minimal drop in accuracy. The source code is available at \url{https://github.com/Div290/CeeBERT}.
  
\end{abstract}
\section{Introduction}

In recent years, Pre-trained Language Models (PLMs) have demonstrated substantial advancements in enhancing Natural Language Processing (NLP) tasks. These models, such as ELMo \cite{peters1802deep}, BERT \cite{devlin2018bert}, ALBERT \cite{lan2019albert}, GPT \cite{radford2019language}, XLNet \cite{yang2019xlnet}, and RoBERTa \cite{liu2019roberta}, encapsulate extensive knowledge transferable to diverse downstream tasks. Despite their remarkable efficacy, large-scale PLMs suffer from inference latencies attributed to their substantial size. With millions or even billions of parameters, these models become computationally expensive, inefficient in terms of memory consumption, and exhibit latency challenges.

Moreover, past approaches like \cite{fan2019reducing, michel2019sixteen, zhou2020bert} have highlighted the concern that PLMs tend to be over-parameterized, leading to the `overthinking' issue. This problem arises when shallow representations at early layers are often adequate for accurate predictions on many input samples, while the final layer's representations may suffer distractions due to over-complicated or irrelevant features that lack generalization. Overthinking in PLMs wastes the computation, degrades model performance and hinders model generalization.

To circumvent this, many variants of the BERT, like DeeBERT \cite{xin2020deebert}, ElasticBERT \cite{liu2021towards}, FastBERT \cite{liu2020fastbert}, PABEE \cite{zhou2020bert}, etc. facilitate inference at the intermediary layers of PLMs through early exits.
In this setup, each sample must ascertain whether the inference can be completed at intermediary layers
or the last layer. The decisions of early exit are based on the confidence at the intermediary layers being above a threshold. 
In the following, we consider the cost quantified as inference time (latency). However, depending on the application, the cost can also be present as other factors like power and computational resources. 

The threshold used to compare the confidence levels significantly impacts the amount of latency and accuracy: with a lower threshold, more samples exit early, but with a lower confidence value, leading to lower accuracy and lower latency. With a higher threshold, fewer samples exit early, leading to higher latency but improved accuracy. Hence, one has to set the threshold that optimally trades-off between latency and accuracy.

\begin{figure}
    \centering
    \includegraphics[scale = 0.47]{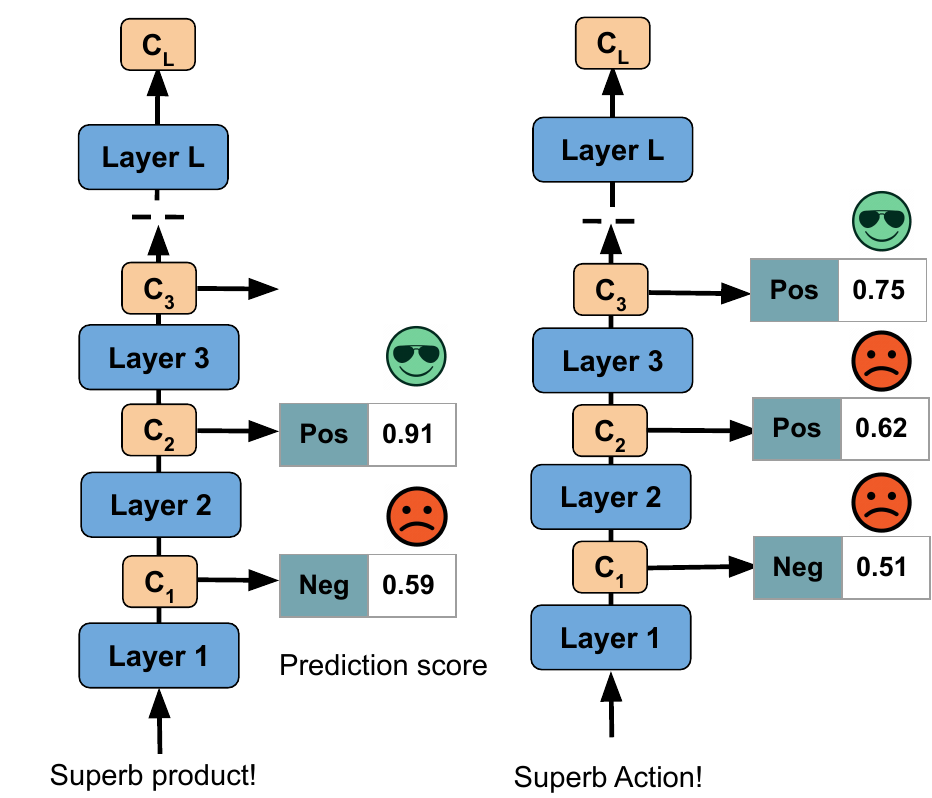}
    \caption{Depiction of cross-domain inference setup for early exit models. The backbone is trained on the source dataset. i) \textbf{Left:} Inference on the source dataset. ii) \textbf{Right:} Inference on target dataset from a different domain. The changes in confidence value is due to change in target domain distribution.}
    \label{fig:main_setup}
\end{figure}

The threshold is often determined using a labeled dataset \cite{xin2020deebert, liu2021elasticbert, schuster2021consistent} or by using some fraction of training data \cite{schwartz2020right, huang2017multi, yang2020resolution, han2023dynamic} during training and serves as a crucial reference point for decision-making during inference. However, a significant challenge arises when deploying pre-trained models that are later tested on samples whose latent distribution can be different from the training samples in a zero-shot setting \cite{pushp2017train, wang2023bert}. For instance, a sentiment analysis model pre-trained on electronic product reviews (source data) to analyze sentiment in a distinct domain, like movie reviews (target data), poses a challenge. The language and sentiment expressions in movie-related content may significantly differ from electronic reviews as illustrated in Fig~ \ref{fig:main_setup}.

PLMs exhibit strong generalization across domains on datasets with similar task types \cite{wang2023bert}. Nevertheless, the distribution of confidence in the prediction of classifiers attached to the intermediary layers can change when transitioning between domains. This can render early exit methods less efficient if the thresholds are not adjusted as per the latent distribution of input samples. This real-world challenge prompts the question: How to adapt the threshold in deployed early exit PLMs to maintain efficiency and robustness to domain shifts in a zero-shot setting? Also, in post-deployment scenarios, samples are fed sequentially, in an online fashion, where inference for each sample needs to be performed before the next sample is fed, this leads to the challenge of learning the optimal threshold in an online and unsupervised manner. To tackle this issue, we introduce an online learning algorithm using the Multi-Armed Bandit framework. \cite{ML02_UCB1_Auer}.

%If the threshold value is large, most of the samples will be forced to pass through the last layer leading to better accuracy but poor latency. If the threshold is small, most of the samples will exit early leading to poor accuracy and also lesser latency costs. 
Our algorithm, \underline{C}ross Domain Inference in \underline{E}arly \underline{E}xits \underline{BERT} (\algo), learns a threshold from a set of thresholds that achieves optimal trade-off between accuracy and latency. We extensively evaluate the performance of \algo{} on five datasets viz. IMDB, MRPC, SciTail, SNLI, Yelp, and QQP cover different classification tasks --sentiment, entailment, and natural language inference.
In our evaluation, we trained the PLM with exits on a source dataset and assessed its performance on a target dataset which exhibits variation in the latent data distribution from the source dataset but shares a similar task type.

\algo{} achieves speedup ranging from $2\times$ to $3.5\times$ in inference time,  while maintaining a minimal accuracy loss of $0.1\%$ to $3\%$ compared to the naive BERT classification. 
Notably, instead of using pre-fixed thresholds learned on the source dataset, CeeBERT learns the thresholds on the fly for incoming samples from the target dataset.  This sets it apart from previous early exit PLMs that rely on validation datasets representative of the target dataset to determine threshold values.

Our primary contributions are as follows:
\begin{itemize}
    \item We introduce cross-domain adaptation in Early Exit PLMs where the thresholds of exit classifiers are adapted without requiring any validation dataset from the target task as used in the previous baselines (see Section \ref{sec: baselines}), thus making Early Exit PLMs robust to domain changes and enhancing their efficiency.
 % \item We dynamically select optimal threshold values based on latent data distribution in an online and unsupervised setup, enhancing the efficiency of early exit PLMs and making them robust to domain changes.
\item In Section~\ref{sec:Algorithm}, we introduce \algo, an upper confidence-based algorithm that relies solely on confidence scores for learning the optimal threshold to make exit decisions.
\item In Section~\ref{sec:Exp}, we evaluate \algo{} on five different datasets drawn from diverse tasks to demonstrate its proficiency in improving inference latency compared to the state-of-the-art algorithms.
\end{itemize}

\section{Related work}
Early exit methods are applied for various tasks such as image classification, image captioning and NLP tasks to reduce the computational resources and inference latency. 

\noindent\textbf{Early exits in Image tasks:}
For image classification tasks, BranchyNet \cite{teerapittayanon2016branchynet} uses classification entropy at each attached exit to decide whether to infer the sample at the side branch based on the entropy of prediction.
Shallow-deep\cite{kaya2019shallow} and MSDNet \cite{huang2017multi} improve upon BranchyNet by effectively choosing the thresholds based on the confidence distribution.
Similar architectures  \cite{laskaridis2020spinn, pacheco2021calibration, dai2020epnet} split the NN to be deployed on edge and cloud. SEE \cite{wang2019see} work in service outage scenarios. FlexDNN \cite{fang2020flexdnn} and Edgent \cite{li2019edge} focus mainly on the most appropriate Neural Network (NN) depth. Other works such as Dynexit \cite{wang2019dynexit} focus on deploying the multi-exit NN in hardware. It trains and deploys the NN on Field Programmable Gate Array (FPGA) hardware while Paul \textit{et al.} \cite{kim2020low} explains that implementing a multi-exit NN on FPGA board reduces inference time and energy consumption.
In a parallel vein, the MuE and DeeCap \cite{tang2023you, fei2022deecap} model employs a distinctive approach to apply early exits to the image captioning. DeeCap only applies to to decoder while  MuE applies to the encoder as well as the decoder.

\noindent \textbf{Early Exit in PLMs:} 
Multiple approaches have been proposed to effectively apply early exits to PLMs and solve multiple NLP tasks \cite{bapna2020controlling,elbayad19arxiv, liu2021elasticbert, ACL2020_Deebert, zhou2020bert, he2021magic, banino2021pondernet, balagansky2022palbert, sun2022simple, ji2023early}. DeeBERT~\cite{ACL2020_Deebert}, ElasticBERT~\cite{liu2021elasticbert} and BERxiT \cite{xin2021berxit} are based on the transformer-based \cite{vaswani2017attention} BERT model. BERxiT proposes an efficient fine-tuning strategy for the BERT model with attached exits. DeeBERT is obtained by training the exit points attached before the last module to the BERT backbone separately. In contrast, ElasticBERT is obtained by training all the exit points attached to the BERT backbone jointly. PABEE \cite{zhou2020bert} is another multi-exit model that makes the exit decision based on the stability of the predictions after different exits. LeeBERT \cite{zhu2021leebert} proposed a self-distillation framework that has similar exiting criteria as PABEE. ETFEE \cite{ji2023early} adds an adapter on top of the transformer layers and an (Entangled Frame) ETF classifier to make intermediate exits learn better.

{\bf Multi-Armed Bandits in Early Exit NN:} LEE \cite{ju2021learning}, DEE \cite{ju2021dynamic} and UEE-UCB \cite{hanawal2022unsupervised} leverage the MAB framework to learn the optimal exit in 
early exit NNs. LEE and DEE mainly focus on learning optimal depth in image classification tasks, while UEE-UCB finds optimal depth for NLP tasks employing a pre-trained ElasticBERT \cite{liu2021towards} model. UEE-UCB does not need any labels but works under the assumption that the prediction of the intermediary layers follows strong dominance property \cite{verma2019Unsupervised}. 

Our approach differs from past works as 1) Unlike previous studies, our work is primarily concerned with making unsupervised cross-domain inference efficient in early exit PLMs. 2) Our work is focused on adapting the threshold values based on the underlying distribution of a dataset. 3) We use the Multi-Armed Bandits framework to solve the problem of threshold learning. We compare against different early exiting models in table \ref{tab: results}.
\section{Problem Setup}
We start with a pre-trained PLM like BERT or ALBERT and attach exit classifiers at each layer. In the following, we discuss how the exit layers are trained and used for early inference. 
\subsection{Training exits  classifers}\label{sec: MAB}
Let $\mathcal{D}$ represent the distribution of the source dataset with label class $\mathcal{C}$ used for backbone training. For any input sample $(x, y) \sim \mathcal{D}$ and the $i$th intermediate classifier, the loss is computed as:
\begin{equation}
    \mathcal{L}_i(\theta) = \mathcal{L}_{CE}(f_i(x, \theta), y)
\end{equation}
Here, $f_i(x, \theta)$ is the output of the classifier attached at the $i$th layer, where $\theta$ denotes the set of learnable parameters, and $\mathcal{L}_{CE}$ is the cross-entropy loss. We learn the parameters for all classifiers simultaneously following the approach outlined by Kaya et al. \cite{kaya2019shallow}, with the loss function defined as $\mathcal{L} = \frac{\sum_{i = 1}^{L} i\cdot\mathcal{L}i}{\sum_{i = 1}^{L} i}$, where $L$ is the number of layers in the backbone. This weighted average considers the relative inference cost of each internal classifier. Subsequently, the model is ready for testing on related tasks from different domains.
% We consider classification tasks with a target class $\mathcal{C}$.
% We use a pre-trained ElasticBERT backbone with $l=12$ transformer layers with classification heads attached to the output of specific layers that output scores for the target classes. We convert the scores into probability vectors by attaching a softmax layer. An input in ElasticBERT is processed sequentially through the intermediary layers outputting probability vectors at layers where classification heads are attached. Processing at an intermediary layer can stop and the sample can exit without passing through the following layers. We utilize the information from the side branches to decide if the sample exits at the intermediary level. More details on how we prepare a pre-trained ElasticBERT model could be found in the Appendix \ref{sec: Elasticbert}.

\subsection{Inference on the Target data}\label{sec: latency_cost}
Let $\tilde{\mathcal{D}}$ be the distribution of the target dataset. Consider an intermediary layer $1\leq i<L$. For an input $x \sim \tilde{\mathcal{D}}$, let $\hat{P}_{i}(c)$ denote the estimated probability that $x$ belongs to class $c\in\mathcal{C}$ and $C_{i}$ denote the confidence in the estimate at the $i$th layer. We define $C_{i}$ as maximum of the estimated probability class, i.e., $C_{i}:=\max_{c\in\mathcal{C}} \hat{P}_{i}(c)$. The decision to exit at the $i$th layer is made based on the value of $C_i$. For a given threshold $\alpha$, if $C_{i}\geq\alpha$ the sample $x$ will be assigned a label $\hat{y} = \arg \max_{c\in \mathcal{C}} (\hat{P}_{i}(c))$. In this case, $x$ is not further processed, and \text{ \it exits} with a label $\hat{y}$. If $C_{i}<\alpha$, then the sample is processed to the next layer. If the sample's confidence is below the threshold for all intermediate classifiers then the sample is inferred at the final layer. We denote the cost incurred in moving the sample from the $1$st layer to the $i$th layer as $o_i$. It denotes the latency or computational cost of processing the sample between the layers $1$ and $i$. Since all the layers of transformer-based PLMs require the same amount of computation, we assume the latency cost $o_i \propto i$.

The confidence can be compared against one of the $k$ possible thresholds denoted by set $\mathcal{A}=\{\alpha_1, \alpha_2, \ldots, \alpha_k\}$. The goal is to identify the threshold which provides the best trade-off between loss in accuracy and latency cost for the latent distribution of the target task.

\subsection{Multi-Armed Bandit (MAB) Setup}

In the MAB setup, a decision-maker iteratively selects actions, adapting to an unknown environment. Each action represents a specific choice, and the goal is to learn which actions result in the most favourable outcomes (highest reward) over time. This dynamic learning process, inherent to MAB setups, resonates with the sequential decision-making required in online learning problems, where choices are made based on incoming data. MAB frameworks, guided by exploration and exploitation principles, enable the learning of optimal actions tailored to the latent distribution.

We treat the set of thresholds $\mathcal{A}$ as the set of actions. Following the terminology of MAB, we refer to them as arms. We define $[L] = \{1, 2, \ldots, L\} $. For any arm $\alpha \in \mathcal{A}$, suppose a sample is processed till layer $i$ and exits i.e. $C_j<\alpha$ for $j \in [i-1]$ and $C_i\geq \alpha$, we define the reward as follows:

\begin{equation}\label{eq:reward}
r(\alpha) = (C_i - C_1) - \mu o_i
\end{equation}
where $\mu$ models the trade-off between accuracy and cost and doubles up as a unit converter to bring both confidence and latency in the same units. 
If the sample does not exit till $(L-1)$th layer then it is inferred at the final layer, where the reward is $r(\alpha) = (C_L - C_1) - \mu o_L$.

The reward could be interpreted as follows: The confidence gained while processing the sample from $1$st layer to $i$th layer reduced by the cost incurred in achieving it (latency).  
Then mean reward for arm $\alpha\in \mathcal{A}$ is
\begin{equation}
    \mathbb{E} [r(\alpha)] = \sum_{i = 1}^{L}\mathbb{E}[\Delta C_i - \mu o_i | \text{exit at $i$}]P(i) 
\end{equation}
where $\Delta C_i = C_i - C_1$ and $P(i)$ is the probability that the sample exits from $i$th layer.
Our goal is to find an arm with the highest mean reward. Since labels of the target samples are not available, we depend on the reward of each threshold to learn their performance. Let $\alpha^{*} = \arg \max_{\alpha\in \mathcal{A}}\mathbb{E}[r(\alpha)]$ denote the optimal threshold. 
Consider a policy $\pi$ that selects threshold $\alpha_t \in \mathcal{A}$ in round $t$ based on past observations. We define cumulative regret of $\pi$ over $T$ rounds as
% \begin{equation}
%     R_{t}(\pi) = r(\alpha^{*})-r(\alpha_{t})
% \end{equation}
% We define cumulative regret over $n\in N$ rounds using instantaneous regret and in terms of expected regret.
\begin{equation}
\label{eqn:Regret1}
    R(\pi, T) = \sum_{t = 1}^{T}\mathbb{E}[ r(\alpha^{*})-r(\alpha_{t})],
\end{equation}
where the expectation is with respect to the randomness in the selection of thresholds induced by the past samples. 
A policy $\pi^{*}$ is said to be sub-linear if average cumulative regret vanishes, i.e., $R(\pi^{*},T)/T\rightarrow 0$. Our objective is to develop a policy learning algorithm with a sub-linear regret guarantee.
\section{Algorithm}
\label{sec:Algorithm}
We develop an algorithm named \underline{C}ross Domain Inference in \underline{E}arly \underline{E}xits in \underline{BERT} (CeeBERT). Its pseudo-code is given in algorithm \ref{alg:algorithm}. The inputs to the algorithm are exploration constant $\gamma$ and latency factor $o_i$ for each layer $i$. For the first $|\mathcal{A}|$ samples, the algorithm plays each arm once. In the subsequent rounds, it plays the arm with the highest Upper Confidence Bound (UCB) index denoted as $\beta_t$. UCB indices are obtained by taking the weighted sum of the empirical average of the rewards $Q_t(\alpha)$ and the confidence bonuses with $\gamma$ as the weight factor. 
If $C_i$ at the $i$th layer is larger than $\beta_{t}$ then the sample exits, otherwise, the sample is passed to the next layer in the backbone while adding latency. If the sample does not exit at any intermediate classifier then it is inferred at the final layer. Finally, the algorithm updates the number of pulls ($N(\beta_t)$) and empirical mean ($Q(\beta_t)$) of the played arm. Note that the algorithm is applied directly to the inference (target) dataset.

\begin{algorithm}[H]
        \caption{CeeBERT}
        \begin{algorithmic}[1]
          \State\textbf{Input:} $o_i \text{ } \forall i, \gamma \geq 1$\\
\textbf{Initialize:} Play each threshold once. Observe $r(\alpha)$ and set 
 $Q(\alpha)\leftarrow \mathbf{0}, N(\alpha)\leftarrow \mathbf{1}, \forall \alpha \in \mathcal{A}$.
\For{$t = |\mathcal{A}|+1, |\mathcal{A}|+2, \cdots$}
\State Observe an instance $x_t$ 
\State \hspace{-.35cm} $\beta_t \gets \displaystyle \arg \max_{\alpha\in \mathcal{A}}\left(Q(\alpha)+\gamma\sqrt{\frac{\ln(t)}{N(\alpha)}}\right)$
\For{$i = 1 \textbf{ to } L$}
\State Pass $x_t$ till layer $i$ 
\State Apply threshold $\beta_t$ and observe $C_i$
\If{$C_i\geq \beta_t$ and $i < L$} 
\State  Infer at layer $i$ and exit
    \State $r_{t}(\alpha) \gets C_i - C_1 -o_i$ 
    \State $N_{t}(\alpha) \leftarrow N_{t-1}(\alpha)+1$
    \State $Q_{t}(\alpha) \leftarrow \frac{\sum_{j=1}^{t}r_{j}(\alpha_j)\mathbbm{1}_{\{\alpha_j=\alpha\}}}{N_{t}(\alpha)}$
\State \textbf{break}

\ElsIf{$i = L$}
    \State Process and infer at the last layer.
    \State $r_{t}(\alpha) \gets (C_l-C_p - o)$ 
    \State $ N_{t}(\alpha) \leftarrow N_{t-1}(\alpha)+1$
    \State $Q_{t}(\alpha) \leftarrow \frac{\sum_{j=1}^{t}r_{j}(\alpha_j)\mathbbm{1}_{\{\alpha_j=\alpha\}}}{N_{t}(\alpha)}$
\EndIf
\EndFor
\EndFor
% \State \textbf{Output: } $\alpha^{*}$ \gets $\gamma_{T}$
        \end{algorithmic}
        \label{alg:algorithm}
\end{algorithm}

Following the analysis of UCB1  \cite{ML02_UCB1_Auer}, one can show that the regret of CeeBERT is of $\mathcal{O}\left(\sum_{\alpha \in \mathcal{A}_p\backslash\alpha^{*}}\frac{log(n)}{\Delta_{\alpha}}\right)$
% \begin{multline}
% \leq 8 \sum_{\alpha \in \mathcal{A}-\alpha^{*}}\frac{log(n)}{\Delta_{\alpha}}+
%     (\pi^{2}/3+1)\\
%     \sum_{\alpha \in \mathcal{A}-\alpha^{*}}\Delta_{\alpha}
% \end{multline}
where $\Delta_{\alpha}=r(\alpha^*)- r(\alpha)$ denotes the optimality gap. For completeness, the proof outline is given in the Appendix (see theorem \ref{thm: Theorem 1}). Hence, CeeBERT comes with a sub-linear regret guarantee.
\section{Experiments}
\label{sec:Exp}
\subsection{Datasets}
We utilized most of the GLUE \cite{wang2019glue} datasets as source datasets and the ELUE \cite{liu2021towards} datasets as the target datasets.
We evaluated CeeBERT on five datasets covering four types of classification tasks. The datasets used for evaluation are: 

\textbf{1) IMDb and 2) Yelp }\cite{asghar2016yelp}: IMDb is a movie review classification dataset and Yelp consists of reviews from various domains such as hotels, restaurants etc. The source dataset for these two datasets is the \textbf{SST-2} (Stanford-Sentiment Treebank) dataset which has a sentiment classification task. \textbf{3) SciTail:} is an entailment classification dataset created from multiple questions from science and exams and web sentences. The source data used was \textbf{RTE}(Recognizing Textual Entailment) dataset which is an entailment classification dataset but with a different context.
\begin{table}
\begin{tabular}{|l|l|l|l|}
\hline
\textbf{Tgt data} & \textbf{\#Samples} & \textbf{Src Data} & \textbf{\#Samples} \\ \hline
IMDb             & 25K                & SST-2            & 68K                \\ \hline
Yelp             & 560K               & SST-2            & 68K                \\ \hline
SNLI             & 550K               & MNLI             & 433K               \\ \hline
QQP              & 365K               & MRPC             & 4K                 \\ \hline
SciTail          & 24K                & RTE              & 2.5K               \\ \hline
\end{tabular}
\caption{This table provides the sizes of the datasets. Src (Source data) is used to train the model to test on Tgt (target data) to evaluate its generalization.}
\label{tab: dataset}
\end{table}
\textbf{4) SNLI(Stanford Natural Language Inference:)} is a collection of human-written English sentence pairs manually labelled for balanced classification with labels \textit{entailment}, \textit{contradiction} and \textit{neutral}. The source data in this case is \textbf{MNLI}(Multi-Genre Natural Language Inference) which also contains sentence pairs as premise and hypothesis, the task is the same as for SNLI but with more general sentences. 
\textbf{5) QQP(Quora Question Pairs)} is a semantic equivalence classification dataset which contains question pairs from the community question-answering website Quora. In this case, the source dataset is \textbf{MRPC}(Microsoft Research Paraphrase Corpus) dataset which also has a semantic equivalence task of a sentence pair extracted from online news sources. Details about the size of these datasets are in table \ref{tab: dataset}. Observe from the table that the size of the source dataset is much smaller as compared to the size of the corresponding target dataset.

\subsection{Baselines}\label{sec: baselines}
We compare our model against three types of baselines:

\noindent\textbf{1) Backbone models:} We choose BERT-base and ALBERT-base as the backbone models which have almost similar performance\footnote{Our method works on any transformer-based PLMs}.

\noindent\textbf{2) Reducing layers:} We directly reduce computation layers, experimenting with the initial $6$ and $9$ layers of the original (AL)BERT model, denoted as (AL)BERT-6L and (AL)BERT-9L. These serve as baselines, setting the lower limit for techniques without additional modifications.

\noindent\textbf{3) Early-exit models:} \textbf{DeeBERT} \cite{xin2020deebert} and 2) \textbf{ElasticBERT} \cite{liu2021towards} employ fixed confidence thresholds for early exits. 3) \textbf{FastBERT} \cite{liu2020fastbert} utilizes a self-distillation framework to train the intermediate exits. 4) \textbf{PABEE} \cite{zhou2020bert} and 5) \textbf{LeeBERT} \cite{zhu2021leebert} uses prediction stability to decide early exits, LeeBERT also distils the knowledge from deeper layers. 6) \textbf{MuE} \cite{tang2023you} relies on hidden representation similarity for early exit decisions and is applied to the BERT-base model for comparative analysis, originally designed for image captioning tasks. 7) \textbf{ETFEE} \cite{ji2023early} and 8) \textbf{PALBERT} \cite{balagansky2022palbert}, state-of-the-art methods, face challenges to adapt to different domains due to bias towards the training dataset. PALBERT uses Lambda layers as explained on \cite{banino2021pondernet}, and ETFEE has an adapter on top of intermediate layers, both contributing to strong bias. Test data must be representative of the training dataset for these baselines.

All the baselines learn the threshold values to decide to exit using a validation dataset representative of the training dataset. Other 
hyperparameters for these baseline models remain consistent with their original implementations, and when applied to the target dataset, we use the same hyperparameters learned on the source dataset.

\begin{table*}[]
\centering
\small
\begin{tabular}{c|cccccccccc}
\hline
Model/Data  & \multicolumn{2}{c}{SST-IMDb} & \multicolumn{2}{c}{SST-Yelp} & \multicolumn{2}{c}{MRPC-SciTail} & \multicolumn{2}{c}{MNLI-SNLI} & \multicolumn{2}{c}{RTE-QQP} \\ \hline
            & Acc          & Speed         & Acc          & Speed         & Acc            & Speed           & Acc           & Speed         & Acc          & Speed        \\ \hline
BERT        & 83.3         & 1.00x             & 77.8         & 1.00x             & 79.1           & 1.00x              & 80.2          & 1.00x             & 71.5         & 1.00x            \\
BERT-6L     & -3.1         & 2.00x             & -3.0           & 2.00x             & -1.6           & 2.00x               & -1.9          & 2.00x             & -1.5         & 2.00x            \\
BERT-9L     & -2.6         & 1.33x          & -2.8         & 1.33x          & -0.9           & 1.33x            & -1.3          & 1.33x          & -1.2         & 1.33x         \\ \hline
DeeBERT     & -2.9         & 2.31x          & -3.5         & 2.13x          & -0.5           & 1.21x            & -2.5          & 2.11x          & -0.9         & 1.35x         \\
ElasticBERT & -2.6         & 2.51x          & -3.2         & 2.95x          & 0.0              & 1.49x            & -1.3          & 2.32x          & -0.3         & 1.80x          \\
PABEE       & -2.4         & 2.42x          & -2.9         & 2.56x          & -0.9           & 1.51x            & -1.1          & 2.39x          & -0.2         & 1.91x         \\
FastBERT    & -2.5         & 2.55x          & -2.8         & 2.62x          & -0.6           & 1.59x            & -1.3          & 2.45x          & -0.5         & 1.96x         \\
MuE    & -2.8         & 2.58x          & -3.3         & 2.81x          & -1.1           & 1.75x            & -1.6          & 2.52x          & -0.2         & 1.88x         \\
LeeBERT     & \textbf{-2.3}         & 2.40x          & -2.7         & 2.49x          & -0.1           & 1.54x            & -1.0            & 2.37x          & 0.0            & 1.92x         \\
PALBERT     & -2.5         & 2.51x          & \textbf{-2.4}         & 2.39x          & -0.7           & 1.63x            & -1.4            & 2.54x          & -0.1            & 1.85x         \\
ETFEE     & -2.6         & 2.45x          & -2.5         & 2.54x          & -0.6           & 1.67x            & -1.8            & 2.55x          & -0.3            & 1.96x         \\
CeeBERT       & \textbf{-2.3}         & \textbf{2.95x}          & \textbf{-2.4}         & \textbf{3.15x}          & \textbf{+0.2}            & \textbf{1.78x}            & \textbf{-0.8}          & \textbf{2.63x}          & \textbf{+0.1}          & \textbf{2.15x}         \\ \hline
\end{tabular}
\caption{Experimental results (median of 5 runs) of early exit models with BERT backbone on the target datasets with 5 random seeds. The format of datasets on top of the table is (source-target) i.e. the dataset before the hyphen is the source and after the hyphen is the target. The accuracy (Acc) is in $\%$ and speed is the Speedup ratio.}
\label{tab: results}
\end{table*}

\subsection{Experimental setup}

\textbf{i) Training of the backbone on source data:} Initially, we train the backbone on the source dataset. We add a linear output layer after each of the intermediate layers of the pre-trained BERT/ALBERT model. We run the model for $5$ epochs. We perform a grid search over batch size of $\{8, 16, 32\}$ and learning rates of \{1e-5, 2e-5, 3e-5, 4e-5 5e-5\} with Adam \cite{kingma2014adam} optimizer. We apply an early stopping mechanism and select the model with the best performance on the development set. The experiments are conducted on NVIDIA RTX 2070 GPU with an average runtime of $\sim 3$ hours and a maximum run time of $\sim 10$ hours for the MNLI dataset.\\
\textbf{ii) Adapting thresholds using CeeBERT:} In this stage, use the model learnt in step (i) to perform inference on the target dataset. In this step, CeeBERT is utilized to dynamically learn optimal thresholds in an unsupervised and online manner for the target dataset. This post-deployment step allows the model to autonomously adapt threshold values based on real-time data, enhancing adaptability in inference. This part is also computed on the same GPU with an average runtime of $\sim 1$ hour. We run each experiment $5$ times where each run includes an online feed of randomly reshuffled input samples to CeeBERT. 

\noindent\textbf{Choice of the action set:} The choice of the action set depends on the total number of output classes, denoted as $C$, within a given dataset. To ensure efficiency and avoid redundancy, we observe that any value in the action set below $1/C$ is extraneous. Consequently, we choose ten equidistant values ranging from $1/C$ to $1.0$. For instance, in a binary classification scenario where the worst confidence value is $0.5$, our action set becomes $\mathcal{A} = \{0.55, 0.6, 0.65, \ldots, 0.95, 1.0\}$. \\
\noindent\textbf{Latency cost and $\mu$:} Recall from section \ref{sec: latency_cost}, we have $o_i \propto i$ i.e. the latency cost for each layer is directly proportional to the depth of the layer in the network. Hence we set $o_i = \lambda i$ where $\lambda$ is the per-layer processing cost. The value of $\lambda$ is user-defined and we set it to $1/L$ so that it is directly comparable to the confidence values. 
The parameter $\mu$ used to model the trade-off between accuracy and latency is set to 0.5. The value of $\mu$ should be set between $\mu \in [0, 1/o_L]$ based on the user's preferences for improved cost or accuracy. The set of choices of $\mu$ is made such that the factor $\mu o_i$ in the equation \ref{eq:reward} is directly comparable to confidence gain. We analyse the model behaviour on changing $\mu$ in section \ref{sec: speedvsacc}.

To maintain consistency with previous methods, we use the speedup ratio as the metric to asses out model which could be written as: 
$$\frac{\sum_{i = 1}^L L\times n_i}{\sum_{i = 1}^L i\times n_i}$$
where $n_i$ are the number of samples exiting from $i$th layer. This metric could be interpreted as the increase in speed of the model as compared to the naive (AL)BERT model. 

\begin{table*}[]
\centering
\small
\begin{tabular}{c|cccccccc}
\hline
Model/Data  & \multicolumn{2}{c}{SST-IMDb} & \multicolumn{2}{c}{SST-Yelp} & \multicolumn{2}{c}{MRPC-SciTail}\\ \hline
            & Acc          & Speed         & Acc          & Speed         & Acc            & Speed                    \\ \hline
BERT        & 83.3         & 1.00x             & 77.8         & 1.00x             & 79.1           & 1.00x               \\
CeeBERT       & \textbf{-2.3 $\pm$ 0.15}         & \textbf{2.95x $\pm$ 0.008}          & \textbf{-2.4 $\pm$ 0.05}         & \textbf{3.15x $\pm$ 0.003}          & \textbf{+0.2 $\pm$ 0.18}            & \textbf{1.78x $\pm$ 0.001}                  \\ \hline
ALBERT        & 82.7         & 1.00x             & 77.1         & 1.00x             & 80.4           & 1.00x   \\ 
CeeBERT       & \textbf{-2.1 $\pm$ 0.12}         & \textbf{2.89x $\pm$ 0.006}          & \textbf{-1.9 $\pm$ 0.08}         & \textbf{2.71x $\pm$ 0.002}          & \textbf{-0.1 $\pm$ 0.16}            & \textbf{1.81x $\pm$ 0.004}                  \\ \hline

\end{tabular}
\caption{The median and standard deviation values of CeeBERT over 5 runs.}
\label{tab: stability results}
\end{table*}
 
\subsection{Results}
In Tables \ref{tab: results} and \ref{tab: results2}, we provide the main results of this paper. We provide median results over $5$ runs. 
The results make evident that CeeBERT consistently outperforms all previous methods both in terms of accuracy and efficiency, due to its ability to adapt and select different thresholds for tasks from different domains. 

This behaviour aligns with the fact that as the target dataset is from a different domain, there is a change in the semantic mapping. This change in turn differs the confidence distribution at the exit points. Previous baselines do not adapt the threshold based on the distribution of target data, hence all the methods get a hit in terms of efficiency. We observe that models trained with more bias toward the training dataset get higher hits in performance, as seen in the current state-of-the-art methods ETFEE and PALBERT. They experience significant performance degradation. This occurs due to the addition of more complex layers (instead of linear layers) and classifiers to the output of intermediate layers, leading to the requirement of the test set to be representative of the training dataset.

\begin{figure*}
    \centering
    \begin{subfigure}{0.31\textwidth}
        \includegraphics[width=\textwidth]{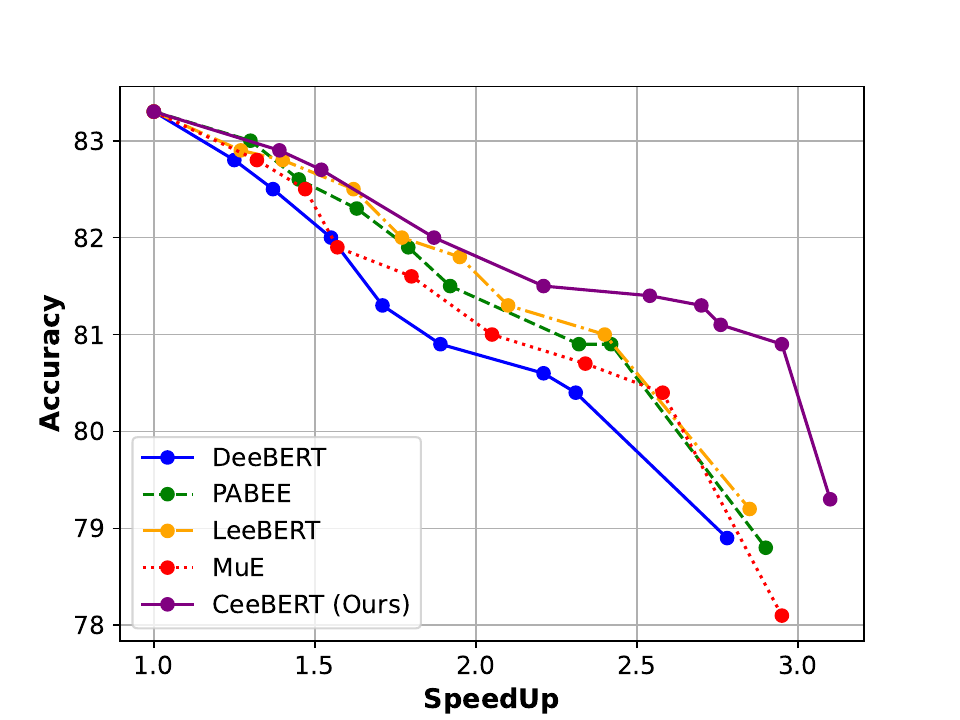}
        \caption{Speedup vs Accuracy curve}
        \label{fig:speedvsacc}
    \end{subfigure}
    \begin{subfigure}{0.31\textwidth}
        \includegraphics[width=\textwidth]{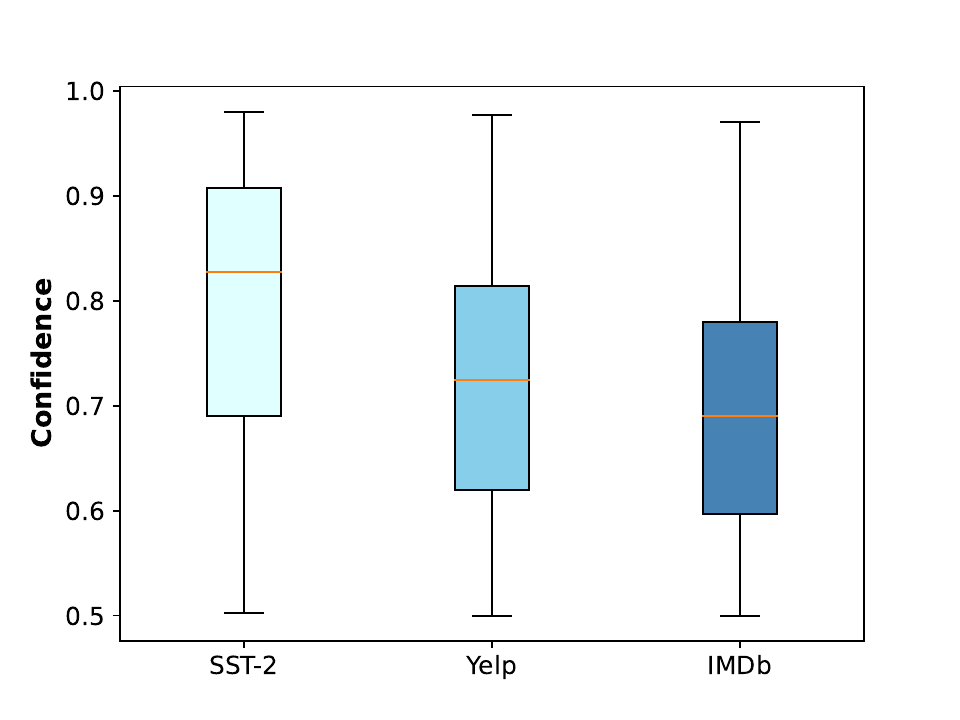}
        \caption{Change in Confidence distribution}
        \label{fig:confidence_dist}
    \end{subfigure}
    \begin{subfigure}{0.31\textwidth}
        \includegraphics[width=\textwidth]{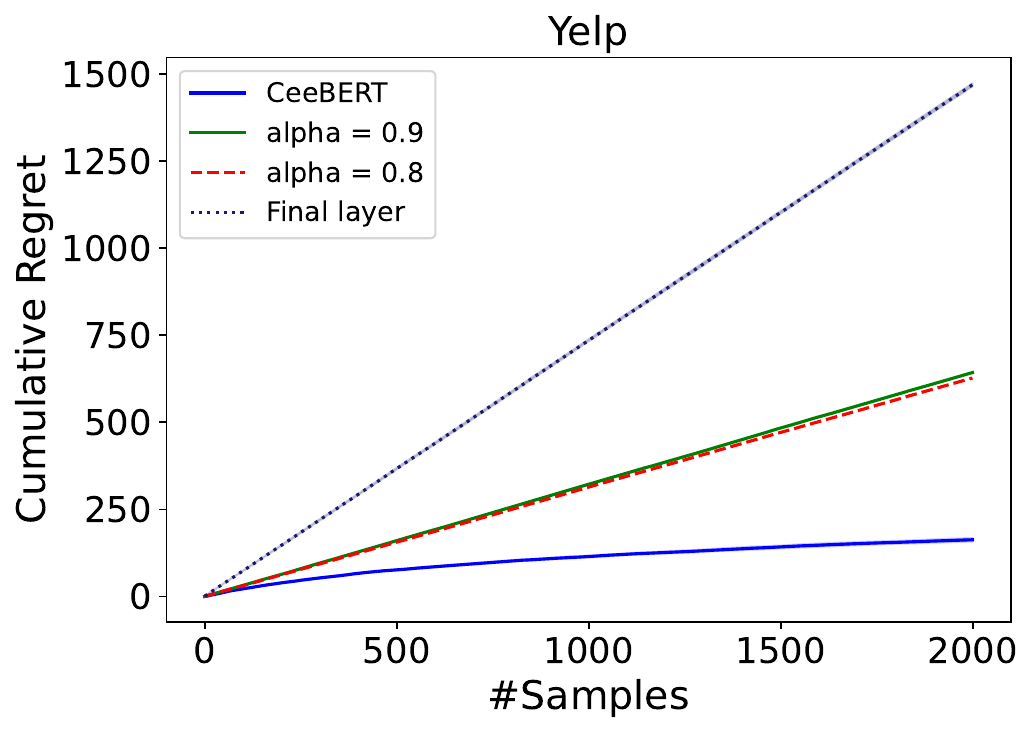}
        \caption{Cumulative regret}
        \label{fig:regret}
    \end{subfigure}
    \caption{\textbf{Left:} We show the trade-off between accuracy and speedup by changing the tunable parameters for various methods. \textbf{Center:} figure states the change in the distribution of confidence at the intermediate exits when the dataset distribution changes. The backbone was trained on SST-2 in this case. \textbf{Right:} figure shows the cumulative regret observed by CeeBERT on the Yelp dataset showing that CeeBERT achieves sub-linear regret.}
    \label{fig:global}
\end{figure*}
CeeBERT's dynamic threshold selection during inference, without requiring full retraining or fine-tuning, offers a substantial advantage. This adaptability speeds up the inference time on average by 2.5x while preserving accuracy. Notably, CeeBERT achieves this without relying on labeled data; instead, it makes real-time threshold decisions based on evolving data distributions. CeeBERT converges to optimal thresholds after just a few thousand samples, as shown in figure \ref{fig:regret}.

The gain on QQP and SciTail datasets as compared to the final layer is explained as the effect of overthinking. Some easy samples get misclassified at the final layer as they suffer from distractions due to over-complicated features available at the final layer. More details on the stability of results i.e. the standard deviation values can be found in the table \ref{tab: stability results} in the Appendix.

\section{Ablation study and Analysis}

\subsection{Accuracy vs Speedup} \label{sec: speedvsacc}
In figure \ref{fig:speedvsacc}, we analyse the behaviour of performance drop over an increase in speedup. DeeBERT and MuE make exit decisions based on confidence values above a pre-defined threshold. PABEE and LeeBERT both use patience-based exiting. By varying the threshold values, we observe variation in the accuracy-latency trade-off. CeeBERT uses the trade-off factor $\mu$ as given in equation \ref{eq:reward}. Increasing $\mu$ will increase the impact of cost in the reward hence it will choose lower thresholds forcing samples to exit early while decreasing it increases the impact of confidence hence improving upon accuracy. We vary these hyperparameters to get the results in figure \ref{fig:speedvsacc}. 

Observe that, adapting thresholds has minimal impact if the confidence is given higher weight in eq.\ref{eq:reward} i.e. $\mu$ is small since it boils down to the case in which confidence is maximized hence many samples are inferred at the final layer similar to other baselines. However, as the impact of cost increases the efficiency of CeeBERT also increases as it forces the samples to exit early reducing overthinking as well as unnecessary computations. It also states that adapting the threshold values while considering both cost and confidence can make early exit models more efficient.

\subsection{Change in confidence distribution}
In figure \ref{fig:confidence_dist}, we plot the confidence distribution of SST-2, IMDb and Yelp datasets when the backbone was trained on SST-2. We plot the boxplots for confidence values for the SST-2 dataset, IMDb and Yelp datasets. The confidence distribution evolves with shifts in the latent distribution of the dataset. Our observations reveal a decrease in the confidence distribution, rendering the threshold learned on the source data less adaptable to the target dataset. This diminished adaptability introduces inefficiencies in the threshold application. The decline in confidence values can be attributed to the distinct semantic structures present in the target dataset compared to the source data. Despite the generalization capabilities of the BERT model to accommodate these changes, a reduction in confidence values becomes evident.

\subsection{Regret Performance}
In figure \ref{fig:regret}, we plot the average cumulative regret on the Yelp dataset over $5$ runs where the samples were randomly reshuffled and fed to the algorithm. This figure gives an idea about the time it takes for CeeBERT to converge for a given dataset. CeeBERT converges to the optimal threshold after exploring on few thousand samples. The plot also contains the cumulative regret when all the samples exited from the final layer and when the thresholds were not adapted and were fixed (same as previous methods). For fixed thresholds, we use $\alpha = 0.5, 0.8, 0.9$. This indicates the importance of learning the threshold values instead of fixing them. For other datasets regret curves refer to the Appendix (figure \ref{fig:regret_curves_appendix}).

\subsection{Stability of CeeBERT}
In table \ref{tab: stability results}, we provide the standard deviation values of 5 runs of CeeBERT on different datasets. These results state the stability of the algorithm. CeeBERT has good stability as it converges fast within the first thousand samples and then the exploration phase is over, now it exploits the optimal threshold to which it has converged. As observed, CeeBERT's stability increases with large datasets as on Yelp the standard deviation values are lower as compared to other datasets. On larger datasets, the fraction of samples over which CeeBERT explores becomes small which in turn increases the stability of the algorithm. The stability is consistent across different backbone models (BERT and ALBERT).

\begin{table}[]
\centering
\small
\begin{tabular}{c|cccc}
\hline
Model/Data  & \multicolumn{2}{c}{SST-IMDb} & \multicolumn{2}{c}{SST-Yelp} \\ \hline
            & Acc          & Speed         & Acc          & Speed         \\ \hline
ALBERT        & 82.7         & 1.00x             & 77.3         & 1.00x             \\
ALBERT-6L     & -3.5         & 2.00x             & -3.1           & 2.00x             \\
ALBERT-9L     & -2.8         & 1.33x          & -2.4         & 1.33x          \\ \hline
DeeBERT     & -3.2         & 2.39x          & -2.9         & 2.24x          \\
ElasticBERT & -2.7         & 2.47x          & -2.6         & 2.41x          \\
PABEE       & -2.4         & 2.42x          & -2.3         & 2.35x          \\
FastBERT    & -2.3         & 2.55x          & -2.5         & 2.32x          \\
MuE    & -2.5         & 2.61x          & -2.8         & 2.51x               \\
LeeBERT     & -2.2         & 2.21x          & \textbf{-1.9}         & 2.49x          \\
PALBERT     & -2.7         & 2.47x          & -2.6         & 2.53x          \\
ETFEE     & -2.8         & 2.65x          & -2.9         & 2.62x          \\
CeeBERT       & \textbf{-2.1}       & \textbf{2.89x}         & \textbf{-1.9}        & \textbf{2.71x}          \\ \hline
\end{tabular}
\caption{Experimental results (median of 5 runs) of early exit models applied to ALBERT backbone.}
\label{tab: results2}
\end{table}
\section{Conclusion}
In this work, we introduced the concept of cross-domain inference in Early exiting PLMs. We proposed an online learning algorithm, CeeBERT that enhances the efficiency of early exit PLMs by adjusting threshold values as per the latent distribution of the incoming data. This adaptation ensures robust performance even in the face of dataset variations within a specific task but across diverse domains. The resulting robustness mitigates the need for any further fine-tuning that reduces time as well as resources. 

% This work contributes to the advancement of adaptable and resource-efficient language models, offering promising implications for real-world applications.

\section{Limitations}

In this work, we have used the same threshold for each exit point. However, one can extend this to having different thresholds at different exit points. However, it comes at the cost of the added complexity of having separate action sets for different layers. It would be interesting to explore and reduce the complexities and further reduce the latency as then the choice of threshold will be made separately for each exit. CeeBERT could be applied to any early exit NN with minor modifications, however, in the experiments, we apply it on (AL)BERT models as done by previous methods.

\section*{Acknowledgements}
Divya Jyoti Bajpai is supported by the Prime Minister’s Research Fellowship (PMRF). 
Manjesh K. Hanawal thanks funding support from SERB, Govt. of India, through the Core Research Grant (CRG/2022/008807) and MATRICS grant (MTR/2021/000645), and DST-Inria Targeted Programme.
\bibliography{latex/custom}
\newpage
\appendix
\section{Appendix}
\label{appendix}
\begin{figure*}
    \centering
    \includegraphics[scale = 0.25]{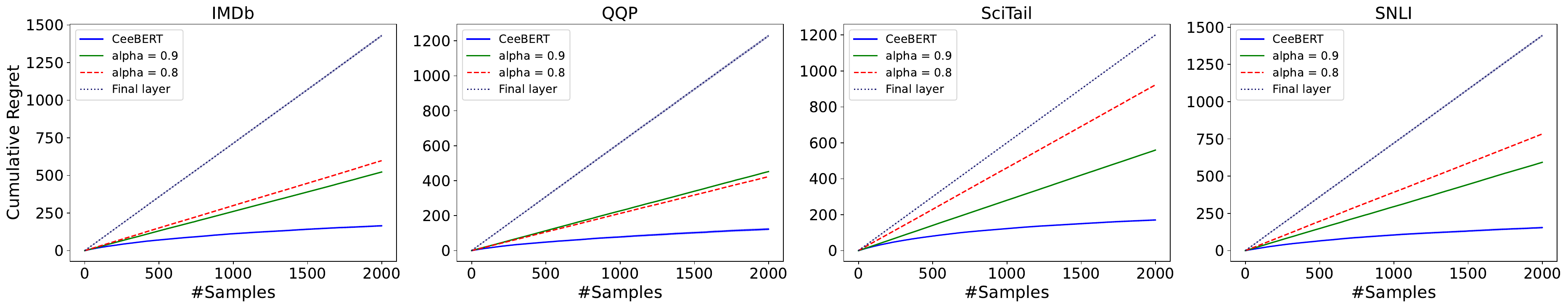}
    \caption{Cumulative regret curves for different datasets.}
    \label{fig:regret_curves_appendix}
\end{figure*}

\subsection{Upper bound on regret of CeeBERT}
\begin{theorem}\label{thm: Theorem 1}For any $\gamma>1$, the regret of CeeBERT with $K$ arms in the action set after $n$ rounds is given as:
\begin{multline}\label{eq: regret bound}
    R(CeeBERT, n)\leq 4\gamma\sum_{\alpha\neq \alpha^{*}}\frac{log(n)}{\Delta_{\alpha}}\\+(\pi^2/3+1)\sum_{\alpha\neq \alpha^{*}}\Delta_\alpha
\end{multline}
where $\Delta_\alpha = r(\alpha^{*})-r(\alpha)$
\end{theorem}

\textbf{Proof: } The proof is very similar to the classical UCB1 \cite{auer2002finite} and follows the same lines with noting the regret in round $t$ as 
$$R_t = r(\alpha_t)-r(\alpha^{*})$$
$r(\alpha)$ is a bounded quantity by definition and more specifically $r(\alpha)\in[-1-\lambda L, 1]$, where $\lambda L$ is the latency cost of the final exit, $L$ is the number of layers and $\lambda$ is the processing cost.

\subsection{Regret performance}
In the case of online learning algorithms, the regret observed is used to monitor the learning process of the algorithm. We observe that across all the datasets, CeeBERT only requires a few thousand samples to converge and from the figure \ref{fig:regret_curves_appendix}, we can also observe that the observed regret is sub-linear as proved on theorem \ref{thm: Theorem 1}. Since these plots are the average of $5$ random runs of the algorithm, we also plot the standard deviation observed, however, the standard deviation plot is barely visible as it is very small and the range of the y-axis is large. We also plot cumulative regret observed when we fix the threshold values as used by previous baselines. 

\end{document}